# Design of an Autonomous Agriculture Robot for Real Time Weed Detection using CNN.

Dhruv Patel*, Meet Gandhi*, Shankaranarayanan H.* and Anand D. Darji.

Sardar Vallabhbhai National Institute of Technology, Surat 395007, India.

**Abstract.** Agriculture has always remained as an integral part of the world. As the human population keeps on rising, the demand for food also increases and so is the dependency on the agriculture industry. But in today's scenario because of low yield,less rainfall, etc., a dearth of manpower is created in this agricultural sector and people are moving to live in the cities, and villages are becoming more and more urbanized. On the other hand, the field of robotics has seen tremendous development in the past few years. The concepts like Deep Learning (DL), Artificial Intelligence (AI), Machine Learning (ML) are being incorporated with robotics to create autonomous systems for various sectors like automotive, agriculture, assembly line management, etc. Deploying such autonomous systems in the agricultural sector help in many aspects like reducing manpower, better yield, and nutritional quality of crops. So, in this paper the system design of an autonomous agricultural robot which primarily focuses on weed detection is described. A modified deep learning model for the purpose of weed detection is also proposed. The primary objective of this robot is the detection of weed on a real-time basis without any human involvement but it can also be extended to design robots in various other applications involved in farming like weed removal, plowing, harvesting, etc. in turn making farming industry more efficient. The source code and other paper related documents can be found at https://github.com/Dhruv2012/Autonomous-Farm-Robot

## 1 Introduction

From the dawn of the human civilization, agriculture has played a pivotal role in fulfilling the demand for the food. With the human population on the rise, the demand for the food has also increased and so is the dependency on the agriculture industry. But the present situation suggests otherwise. There is a dearth of manpower created in the agricultural sector and people are moving to live in the cities, and villages are becoming more and more urbanized because of the irregularities in the rainfall, less yield, etc. At the current growth rate of the world population, it is really important to address the issue of the dwindling agricultural sector. The field of Robotics on the other hand evolved exponentially during the past few decades. Robotics deals with the construction, design, operation, and use of robots, and apart from that it also deals with their computer systems for the processing of information and for obtaining the sensory and control feedback. The emerging area of applications for the robots or drones in the agriculture are weed-control, seed planting, soil analysis, environmental monitoring, and harvesting [3,6,10].

---

* Denotes equal contribution

## 1.1 ROS for Agriculture

ROS stands for Robotics Operating System. It is a framework developed for writing the independent robot software. It consists different compilation of libraries, tools, and conventions which is used for streamlining the process of developing sturdy and complex robot behavior across an ample range of robotic platforms. For incorporating ROS to automate the agriculture, it can be accomplished by making the present agriculture machines ROS-compatible, and/or by introducing ROS-compatible robots into agriculture. To make a ROS-compatible agriculture robot, certain steps are involved and they are; developing the description model of robot components along with the sensor's drivers and controller interfaces. Apart from this, it also involves creating certain specialized ROS packages for the robot in order to make it perform the required agriculture tasks. A similar kind of process is followed when the ROS compatible machinery are being used into agriculture. The ROS-compatible robots can use the vibrant features of ROS and other open-source technologies in order to obtain better results once the process of incorporation of ROS packages are completed [7].

## 1.2 Deep Learning and Computer Vision for Agriculture

Deep Learning is a subset of Artificial Intelligence (AI) where the models mimic the neural connections of a human brain for recognizing patterns in the data which can be used for decision-making [1]. With the availability of adequate training data and increasing use of deep neural networks, the hardware side has also been developed to provide high computing power for these algorithms. This has opened a wide range of possibilities, and it has certainly impacted the agriculture field too. A variety of computer vision and Deep Learning-based approaches are being utilized for extracting meaningful phenotyping information from crops. These approaches helped us to develop a novel solution for weed control through AGRIBOT.

For this paper, we have simulated an agricultural robot called "AGRIBOT". AGRIBOT - Autonomous Agricultural Robot is a four-wheel skid steering [13] prototype that is designed to be used in the agrarian environment for automating different tasks like monitoring and classification between crop and weed. AGRIBOT is equipped with onboard sensors including Global Positioning System (GPS), Compass, Inertial Measurement Unit (IMU) and Camera.

## 2 Literature Survey

Traditional Weed Management required methods like burying, cutting, or uprooting. Several tools like finger weeders, brush weeders etc., were used which performed the above-mentioned operations. But these methods required human intervention, which makes them less effective, time taking & it caused damage to crops in case of intra-crop weeding. This led to the demand for robotics and automation in the field of agriculture. Various Robots have been developed which navigated autonomously and performed various tasks like plowing, crop monitoring, seeding through actuation in real-time. The recent breakthroughs in the domain of Modelling and simulation, sensor fusion, crop-weed classification with the help of computer vision, deep learning and how they provided the base for the development of AGRIBOT is thoroughly discussed in this section.

### 2.1 Sensor Fusion

It is difficult for a mobile robot to estimate its exact location even if it has been equipped with a map of a specific environment. In addition, the exploration of an unknown environment of the robot's location is more difficult. These difficulties are mainly caused by the accumulative error of the measuring sensors, the dynamics of robots, and environmental changes.

Authors have tested navigation algorithms using single GPS receiver and multi receivers. When using just one single GPS receiver, the reception impediment and noise resulted in the serious inaccuracy of positioning. Therefore, they used three GPS receivers to make the positioning of GPS is more accurate and reliable. They have also used camera sensor to compensate for errors. Fused data has been applied to compensate the tracking error between the real path and the target path through the Kalman filter. Though the robot traced the path efficiently, there was an error in co-ordinates shifting [9].

Likewise, Magnetometer is also an important sensor in the ground vehicle, unmanned aerial vehicles and even in satellites and submarines. Three-axis magnetometers are mostly used in the present scenario. Constant and time varying sources corrupt the measurements from Magnetometer. The errors which vary through time come from the nearby electronics whereas the constant sources consist of hard irons error, soft irons errors, non-orthogonality and scale factor errors. Hence it been a challenge for researchers to calibrate and compensate these errors. Many algorithms such as moving median, Kalman filtering, unscented Kalman filtering, etc. have been tested and have proven to be successful in removing the noise up to a certain extent [12].

The literatures which were reviewed for the sensor fusion part focused on using only one type of sensor (i.e., either Magnetometer or GPS) but for AGRIBOT, we have explored the possibility of integrating different types of sensors like magnetometer, GPS, IMU and camera in order to bring down the positional errors and trace the path accurately.

## 2.2 Crop-Weed Classification

In earlier times, Robots did not have the capability to detect and classify objects at real-time. Due to development in the field of imaging systems, robots were incorporated with cameras, for example - Colour cameras performed the task of segmentation of plants while the spectral cameras performed the task of classification of plants. Various image processing and Deep Learning algorithms have made it possible to classify multi-weed species and crops under outdoor illuminations [15]. The errors of the algorithm, when processing 666 field images, ranged from 2.1 to 2.9%. The ANN correctly detected 72.6% of crop plants from the identified plants and considered the rest as weeds. It basically uses thresholding to segment the vegetation mask from the image and then the morphological features are given to an ANN network for classification [15].

Due to the rise in the development of CNNs, several architectures have been proposed to detect or classify vegetation masks in images. One of the approaches uses this mask along with bounding boxes to train a VGG 16 network to classify extracted blobs into crops and weeds. It is trained on 1500 images, validated on 350 images, and tested on 150 images, and achieves around 90% accuracy in classifying crops and weeds species [21].

The above bounding box-based approaches suffer from overlapping issues of the boxes. Hence, to counter this and get pixel-level accuracy, several semantic segmentation algorithms are proposed which classify each pixel into a class. As referred from [8], a CNN is proposed based on the same encoder-decoder based segmentation architecture which classifies crop and weed in real-time and can operate at around 20Hz. It is a 14-channel network trained on the Bonn dataset and has less than 30,000 parameters. These multichannel representations along with RGB converge to 95% accuracy which is greater by approximately 9% than the RGB counterpart. The network is capable of running on 5Hz on a flying vehicle [8]. Inspired by this model, we have implemented a 10-channel network for the crop-weed classification task.

## 3 System Design

For the application of controlling the weed which are surrounding the crops in the field, these points must be taken into consideration. The terrain in which the vehicle will be driven is highly uneven. The Wheels should be rugged enough for driving in these conditions. The drive system should be able to take a zero-radius turn if required. The track length of the vehicle may vary. The minimum value of the ground clearance of such a robot should be 200mm so that the on-board modules, as well as crops, might not be damaged. On-board power supply and connectivity must be available to the robot. It should be able to handle the data available on the real-time basis and make decisions based on that in real-time scenario. Precise monitoring and detection of weed are more important so that farmers can easily take care afterward. The main processor should be capable of handling multiple data coming from different inputs at different frequencies.

Four-wheeled systems are more efficient compared to three or two-wheeled systems because they are more stable on the rough surfaces as compared to three-wheeled systems and cornering can be done more efficiently. The four-wheeled robot is pretty much well balanced because the center of gravity is usually in the middle and due to this reason, this system is commonly used in automobiles and robotics [5]. So, a four wheeled system will suit perfectly for an agriculture robot. The overview of AGRIBOT highlighting major systems is given in Fig. 1. The robotic structure consists of a primary processor, secondary controller for actuation, various sensors and actuators using which the farmer can monitor and visualize the field remotely.

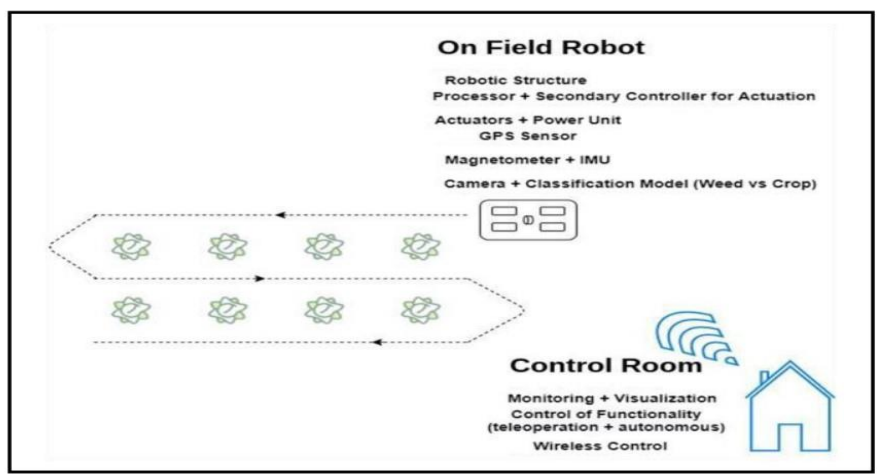

**Fig. 1.** Overview of AGRIBOT.

### 3.1 Proposed Mechanical System Design

With regards to the mechanical structure of the robot, certain design margins like weight and size of the body, strength, manufacturability, time for manufacturing, and cost. The mechanism responsible for driving the robot consists of DC motors, leg structure links, bearings, and shaft at the output which drive the wheels [2,8]. The structure of the chassis is simple and aluminum box section is used for its construction. Design is kept in such a way that the length between sides can be changed according to field requirements. Primarily we have considered the dimensions according to a sugarcane farm.

All the parts have been designed using CAD (Computer Aided Design) tool Solidworks to enable further stress analysis and complex frame structure of the robot. This design comprises of the parts which are to be fabricated and the other parts supplied from off the rack. While designing the robot's body parts, assimilability with other parts such as motors and cameras has been considered. The developed 3D design allows to study and analyze the mechanical interface between each part as joints and links. URDF of the robot has been generated using a custom add-on available in CAD software. Physical criteria of the robot body are also applied through this exporter. Then URDF scripts were used in the ROS framework to simulate the robot for further functionality.

The TF (Transform) tree is used to describe the various frames of the robot and their relative relationships. The TF tree is required for performing the complex robot kinematics operations. With the incorporation of the TF tree with the URDF model of the robot, all operations involving kinematics is performed by ROS's TF package which in turn saves the developers' time for performing these complex mathematics operations manually. We have used two packages from ROS libraries, robot state publisher and joint state publisher. These packages are responsible for all background coordinates transformations.

The kinematic chain generated by the URDF file for this has 8 links and 9 joints. Joints connect multiple links to one link and they can be defined as parent and child links as per the URDF terminology. The joints responsible for driving are defined as continuous. Continuous joints have given to 4 joints so that all wheels can rotate freely around one axis. Dynamic simulation using Gazebo and RVIZ (Robot Visualization) for visualization is possible for this robot definition. On the Robot local frame, this robot has four DOFs overall, which in turn provides three DOFs motion ($V_x$, $V_y$, and $\omega$) in the frame of world. The locomotion control system for this robot is skid steering system.

### 3.2 Proposed Embedded System Design

The proposed embedded system of the agricultural robot can be divided into five parts. They are the central processing unit, control unit, power supply unit, peripheral circuitry, and motors.

The Central Processing Unit includes a primary processor which is used for processing the data available from the GPS Sensor, Camera, and IMU sensor. The primary processor proposed here is NVIDIA Jetson Nano [4].

The Control unit consists of a slave controller, encoder, and motor driver circuit. The secondary controller is responsible for driving the motors and taking the feedback from the motors via encoders. The secondary controller used in the agricultural robot is Arduino Mega 2560[14]. The encoder is used for translating the linear or rotary motion into a digital pulse which in turn would control the direction, speed, position or distance of the motor. The encoder is used to precisely control the displacement of the agricultural robot. The encoder used in this robot is in-built in the motor itself. Motor drivers are used because the motor draw a huge amount of current but the controller works on low currents. These drivers take the low-current control signal from the controller and then turn it into a high-current signal so that the motor can be driven. The primary focus of the agriculture robot is to detect the weed so a low RPM and high torque motor will meet the requirements.

The Peripheral Circuitry has three components present in it. They are Raspberry Pi V2 Camera, MPU 9265 IMU, and Neo-M8N GPS Module. These are directly interfaced with the primary processor (i.e.) using CSI, I2C, and USB Interfaces respectively.

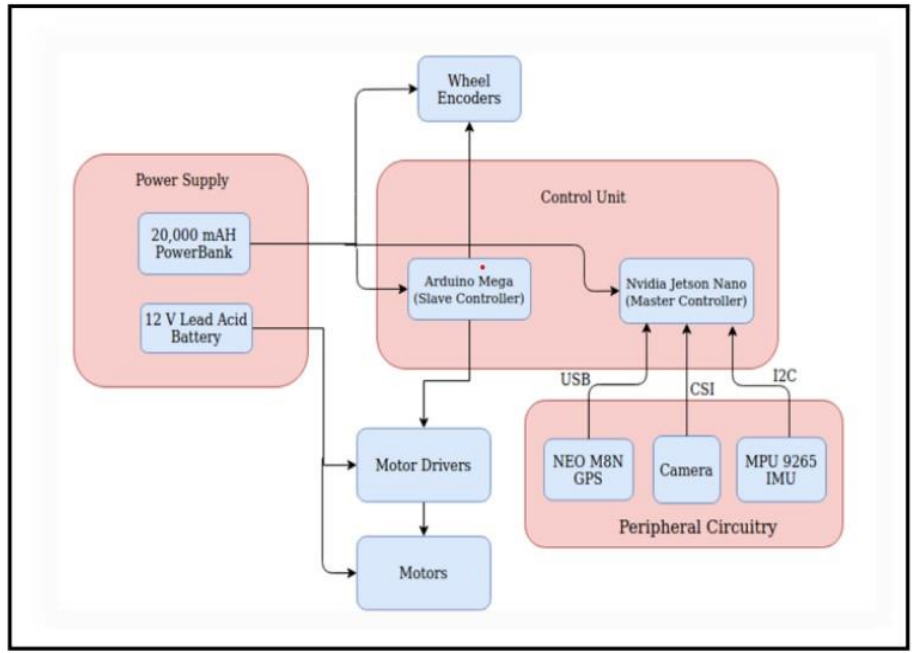

**Fig. 2.** Embedded System Block Diagram of Agricultural Robot.

### 3.3   Modelling and Simulation of AGRIBOT

Gazebo is a multi-robot simulation tool. It is capable of simulating a swarm of robots, sensors and objects in a 3-dimensional world in an accurate and efficient way. It can generate the practical feedback from the sensor and for generating interactions between objects, robots and environment through URDF (Unified Robot Description Format) scripts, it has a robust physics engine. It is available as a stand-alone open-source software, but it is also been offered along with ROS as a simulation tool.

All the physical joints and links, along with the control and sensing interfaces of AGRIBOT are described in this section. The farm environment is modeled and the robot model is simulated in that environment in the gazebo simulator. By including the Open Dynamics Engine – ODE, Gazebo provides the dynamics simulation by taking different physical parameters like friction, contact forces, and gravity into account. Different types of Gazebo plugins have been enabled to develop control interfaces and sensing systems for AGRIBOT. The general architecture and the necessary components in Gazebo for designing a robotic system is described in Fig. 3.

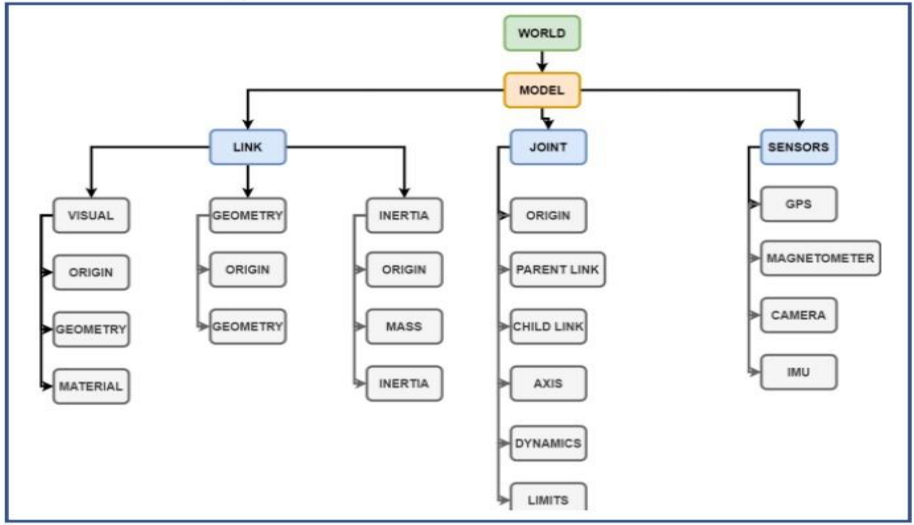

**Fig. 3.** General architecture and the necessary components in Gazebo for modeling a robotic system.

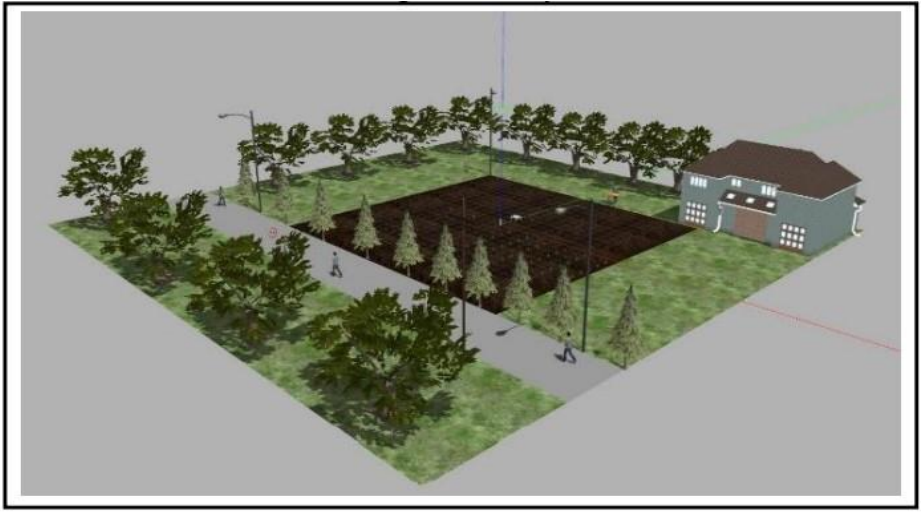

**Fig. 4.** Simulated Environment in Gazebo.

The world is a general term to portray objects, global parameters, and physics properties. Fig. 4 shows a simulated environment in Gazebo. Stationary objects as the modeled field, pine trees, building, street lamps, and farm are incorporated in the world environment. These objects were developed with the help of SDF scripts and models were integrated from GAZEBO library.

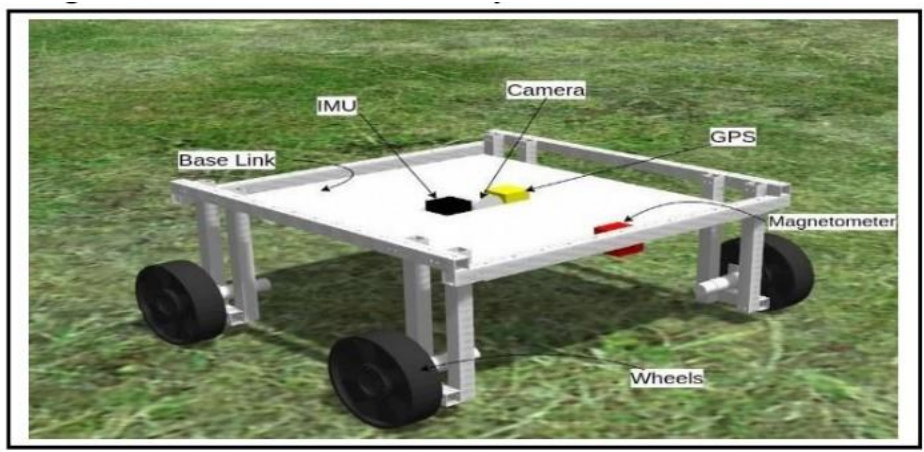

**Fig. 5.** Kinematics Chain of AGRIBOT Platform

The robot in this modeled world is a dynamic object, and it include links which are interconnected to each other with the help of joints. In Gazebo, a link is used to describe the dynamic properties and kinematics of a physical link in the form of inertia information and visual/collision geometry. A joint is used for modeling the kinematic and dynamic properties of a physical joint. This information is described in URDF file format in order to frame the constraints of the design. Fig. 5 depicts the kinematic diagram of the AGRIBOT platform. It shows the kinematic chain of all the physical and virtual links and joints, each joint which possess the information of its placement with regards to the previous link and all these calculations are being done by the TF package.

The robot has different on-board sensors and each of them can be independently simulated as a Gazebo plugin and in the robot model it must be physically attached as a link. In the URDF file, these plugins are incorporated. In Gazebo, the sensor plugins are noiseless and they perceive the environment which has been simulated with perfection. In order to make the plugins more realistic, the noise can be incorporated. For the plugins, a gaussian error of the first-order is used as a noise model and it is added with the measurement taken from each sensor. It is modelled by setting the mean and the standard deviation of the Gaussian distribution. Each measurement of Y(t) at time t is given by:

$$Y(t) = Y' + B + N_Y \quad (1)$$

$$B' = -B/\tau + N_B \quad (2)$$

Y' is the raw measured value, B is the bias/offset, $N_Y$ is additive noise that affects the measurement, and $N_B$ defines the characteristics of random drift with the time constant $\tau$ this includes drift due to environmental changes, voltage changes and age of sensors. [8]

The inertial measurement unit (IMU) reports the robot body's angular rates from a three-axis gyroscope, absolute orientation around the Z-axis by a magnetometer and acceleration from a three-axis accelerometer. With the integration of these measurements with other sensors, a proper reference for the localization of the robot system is obtained.

Through several pre-processing steps, images are being converted into video and DL (Deep Learning) model has been implemented to classify between crop and weed. The necessary parameters like image width and height, frame rate, etc., can be defined within Gazebo plugin through URDF. For each pixel the gaussian noise is sampled individually and it is added to each color channel of that pixel. GPS sensor reports latitude, longitude and altitude of robot body from reference provide through parameters. Integration of this sensor with IMU, after converting it into an XY frame helps us to locate the robot in an outdoor environment.

For controlling the motion in one degree of freedom for each joint, the Gazebo control plugin should be enabled. Apart from this, interfacing of Gazebo with a robot middleware like ROS is also required for controlling each joint. A meta-package called gazebo_ros_pkgs is available which provides different packages for interfacing ROS and GAZEBO. Apart from this, the hardware interfaces, toolboxes, and controllers for controlling the joint actuators is provided by a set of packages called ros_control. The joints are interfaced with the controllers by the hardware interfaces relevant to them. The software control system which has been developed can be used to control the physical robot. The feedback to the control system is read from the encoders. The overview of the control system plugin for the simulated AGRIBOT is portrayed in Fig. 6.

When the ROS package containing the URDF file of AGRIBOT is launched, the simulated model is opened in the virtual world of GAZEBO. The control interface is also loaded and then it waits for all the controllers to be loaded.

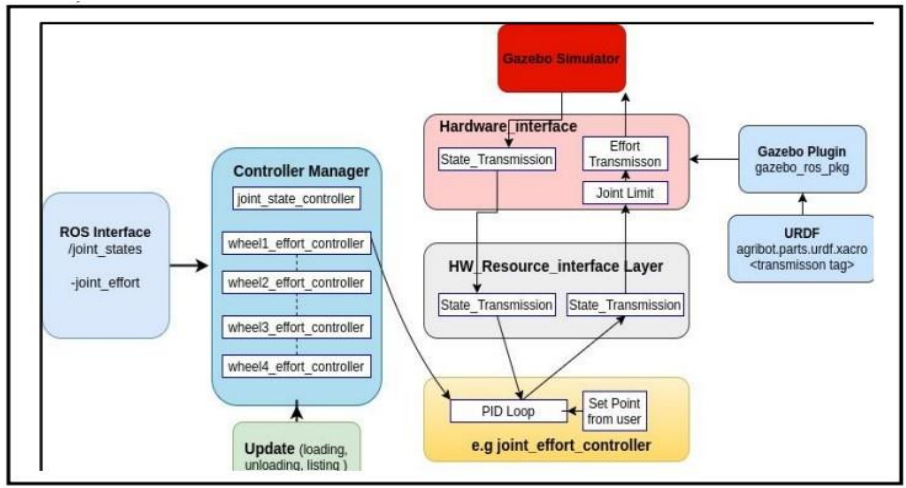

**Fig. 6.** Control system plugin for the simulated AGRIBOT

## 4 Algorithm Implementation: Path Planning and Weed Detection

### 4.1 Path Planning: -

A number of algorithms have been developed for autonomous traversing and monitoring of the outdoor environment in the agricultural domain. Majorly these algorithms consist of Kalman filtering & estimation of robot's position, mapping, path planning, classification algorithm for weed and crops etc. for filtering out the coarse data from the sensors, two filtering techniques namely Moving Median Filter and Single dimension Kalman filter were deployed.

Apart from this, the control and navigation algorithm is responsible for driving the robot with the desired input speed. With the help of inverse kinematics, the information about the desired speed is received in the form of linear, angular velocities and the driving speed for each wheel in the local frame is given as an output by the controller [11]. In order to control the position, velocity, or effort on each joint, these outputs are set points for the relevant ROS controllers. Fig. 7 shows the pictorial view of the path planning algorithm.

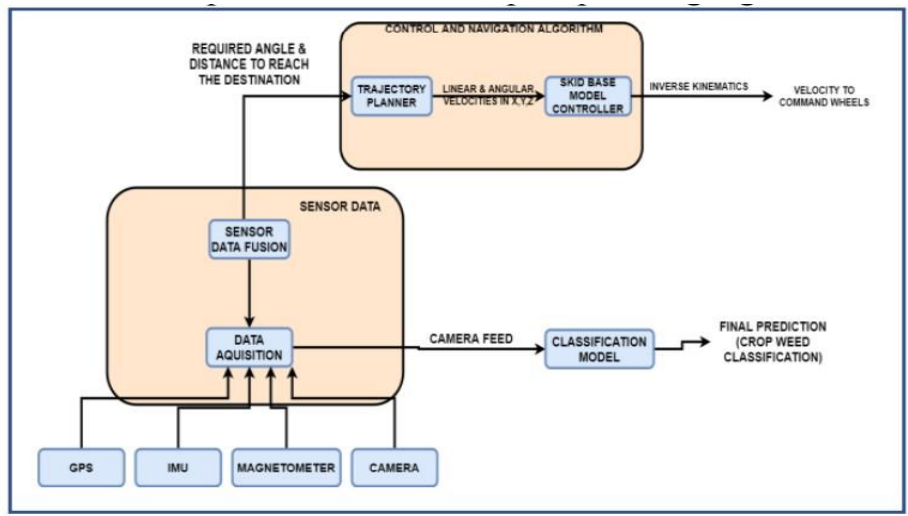

**Fig. 7.** Pictorial View of Path Planning Algorithm.

### 4.2 Crop-Weed Classification

We implemented 2 models – Unet [16] and modified Bonnet, inspired from [17]. The no. of trainable parameters in the UNet model are around 23,54,785 whereas it is 100x lesser in the later one i.e., around 30,000. So, we implemented a 10-channel network based on Bonnet as it is computationally efficient, easier to train and deploy in real-time scenario as compared to UNet [16]. Fig. 8 describes the architecture of the Bonnet Model.

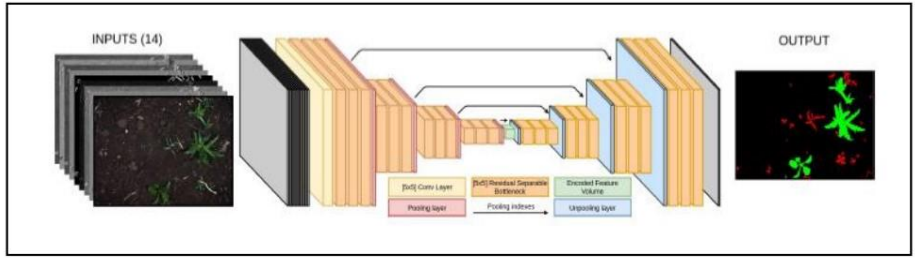

**Fig. 8.** Bonnet Architecture [17]

The input to the network is of 512*384 size. The 10 input channels are namely R, G, B, H, S, V, Excess Green (ExG), Excess Red (ExR), Color Index of Vegetation Extraction (CIVE), Normalized Difference Index (NDI). Along with RGB and HSV color space, these vegetation indices are calculated and supplied as a concatenated representation to normalized input. This aids us in vegetation segmentation (crops, weeds, or soil) and helps to generalize better regardless of the type of crop, outdoor lighting conditions, etc. [17].

Now, the further task is training the network. We tried 2 loss functions to train the network for the semantic segmentation task - Categorical cross-entropy Loss and Weighted Categorical Cross-Entropy Loss (WCCE) [18].

With the Categorical cross-entropy loss, all pixels were given equal weight in the loss optimization irrespective of their class. Hence, it was observed that the network mainly tried to correct its predictions on the major class (crops or soil) in the ground truth. Thus, any false predictions for the minor class(weed) did not get enough weight in the loss function to optimize and correct its prediction. Hence, the above loss function was not effective. Fig. 9 shows the performance of Modified Bonnet Model with Categorical Cross-Entropy Loss on Bonn Dataset.

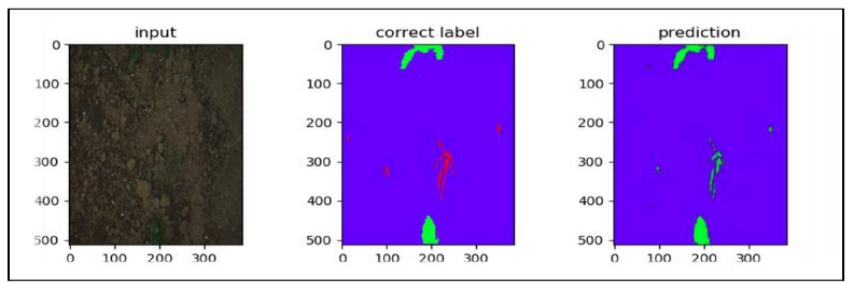

**Fig. 9.** Modified Bonnet's performance with Categorical cross-entropy on Bonn

Thus, for countering the previous issues, we used WCCE loss. Here, the losses of corresponding classes in the ground truth are weighed in inverse proportion to their frequency. Hence, we were able to optimize and correct minor class(weed) as well due to higher weight. It was observed that the model performed better on the class-wise prediction with this loss function. Fig. 10 shows the performance of Modified Bonnet Model with Weighted Categorical Cross-Entropy Loss on Bonn Dataset.

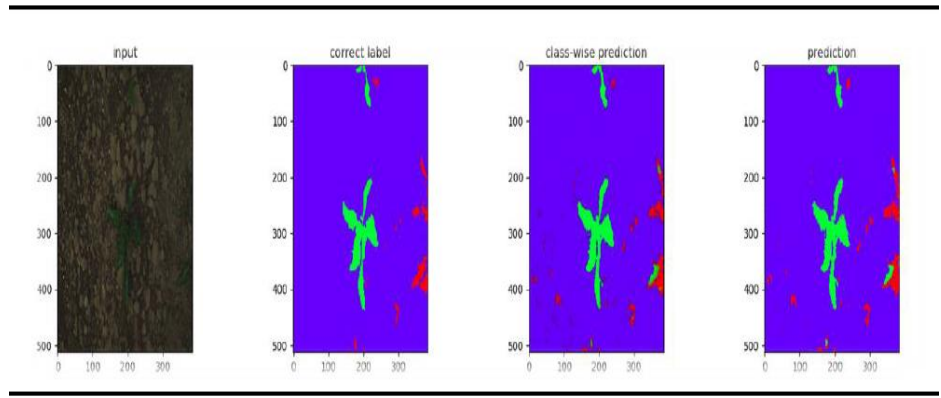

**Fig. 10.** Modified Bonnet's performance trained using WCCE on Bonn

## 5 Results Analysis

This section provides the experimental results and performance of our robot through ROS third-party software and custom open-source tools. For some scenarios, we have given input velocity from the keyboard to ensure the functionality of the robot. The results from the experiment were recorded in a log file for further visualization with live performance, hence graphs were plotted. Fig. 11 shows the magnetometer reading simulated with the help of IMU plugin on Gazebo. Fig. 12 shows the visualization of the simulated camera by showing the image of an example scene in Gazebo.

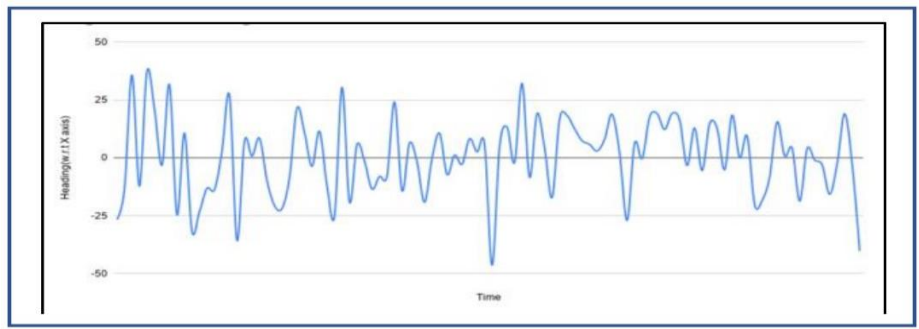

**Fig. 11.** Magnetometer reading simulated with the help of IMU Plugin on Gazebo.

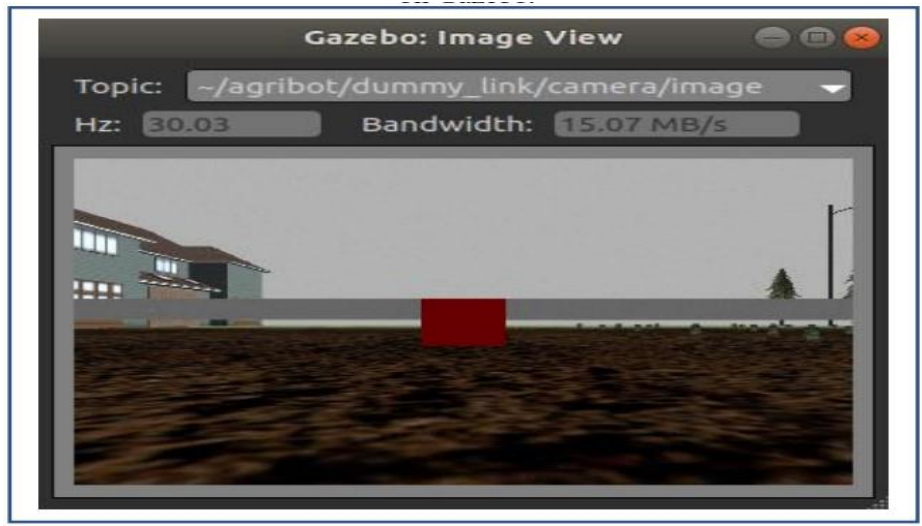

**Fig. 12.** Visualization of simulated camera in Gazebo.

Fig. 13 and Fig. 14 show the results after incorporating the moving median filter and single dimensional kalman filter on the raw magnetometer data respectively. Kalman filter provided better heading correction results as compared to the moving median filter.

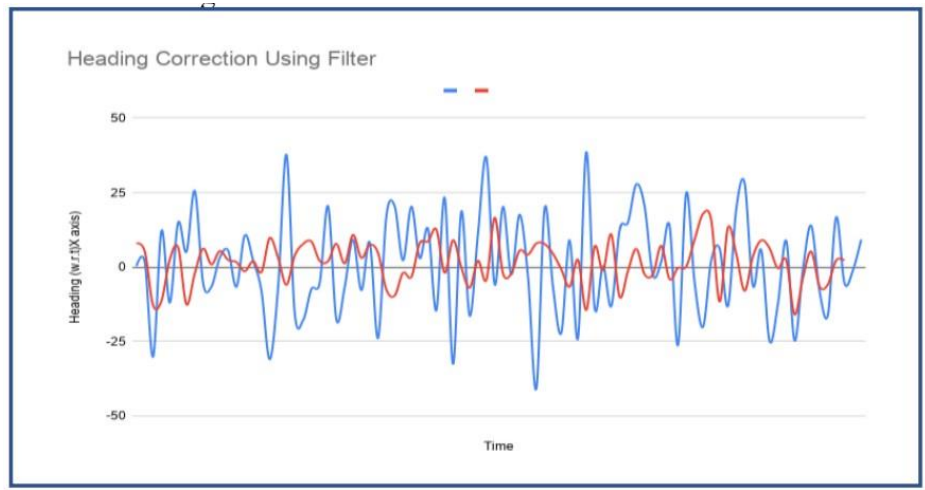

**Fig. 13.** Implementation of moving median filter on the raw data obtained from magnetometer. Blue color represents the raw data and red color represents the filtered data.

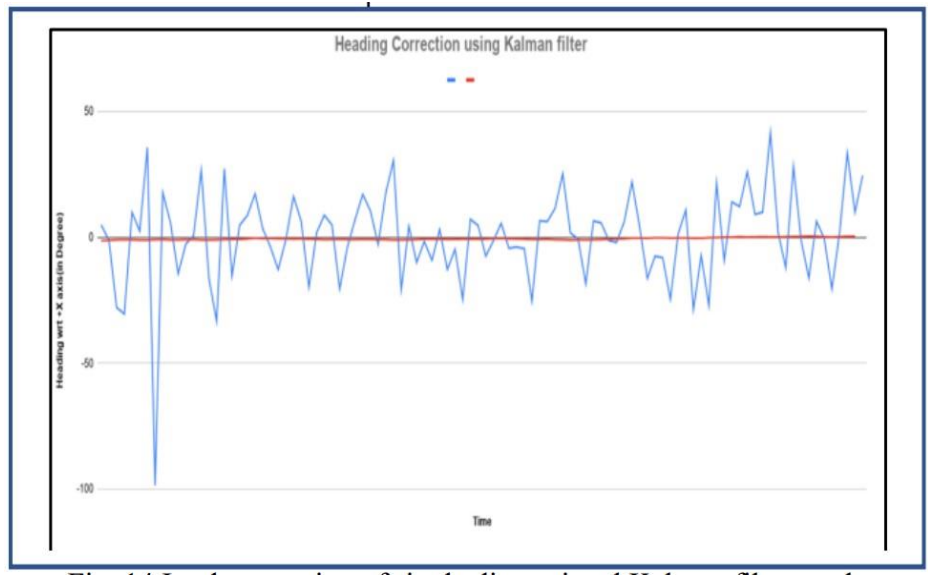

**Fig. 14.** Implementation of single dimensional Kalman filter on the raw data obtained from magnetometer. Blue color represents the raw data and red color represents the filtered data.

In the Fig. 15, the Blue Line is the plotted trajectory, the Black Cluster is the starting point of field and on the left, we have mapviz visualization and on the right, we have the Gazebo simulator. The autonomous traversing through the crops is shown in Fig. 16.

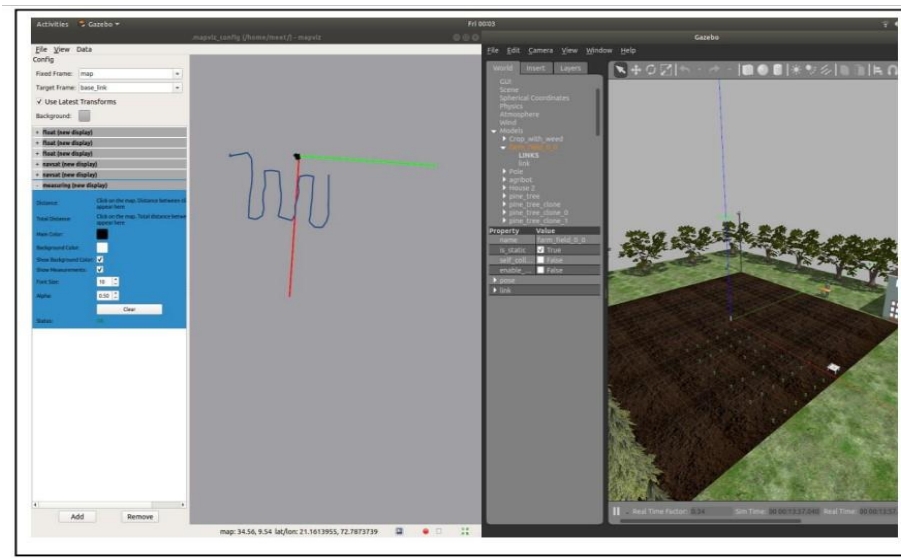

**Fig. 15.** Field traversing through teleoperation.

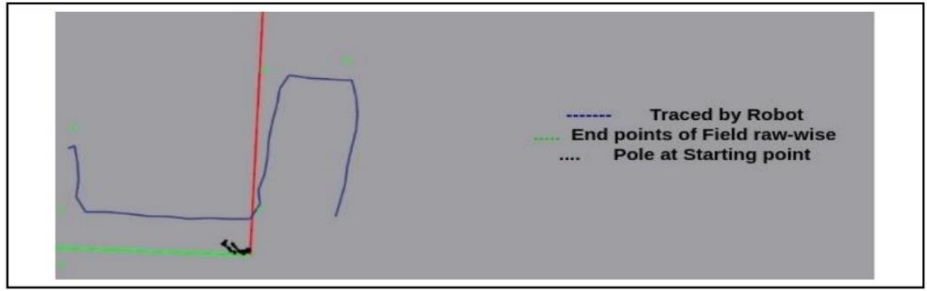

**Fig. 16.** Autonomous Traversing of crop rows in the field.

As mentioned in section 4, we considered only 10 channels instead of all the 14 channels from [17] for the modified bonnet model. Table 1 shows the results obtained by this proposed modification along with the model results from [17] on Bonn Dataset. We have validated our model on these metrics namely accuracy, precision, recall, and mean IoU.

**Table 1.** Performance of Modified Bonnet Model and Bonnet Model from [17] on the Bonn Dataset.

| Model | mAccuracy [%] | mIoU [%] | Precision [%] | | | Recall [%] | | |
|---|---|---|---|---|---|---|---|---|
| | | | Soil | Weed | Crop | Soil | Weed | Crop |
| Bonnet Model [17] | 94.74 | 80.8 | 99.95 | 65.92 | 85.71 | 99.53 | 85.25 | 97.29 |
| Modified Bonnet Model (This Work) | **99.47** | **98.03** | 99.95 | 36.08 | 85.84 | 99.58 | 61.68 | 93.86 |

The accuracy of a model is a measure that describes how it performs across all classes while precision is measured as the ratio between the True Positives and all the Positives. The recall is the measure of our model correctly identifying True Positives. The amount of overlap between the predicted and ground truth bounding box is specified by the intersection over union (IoU) value, which ranges from 0 to 1. Table 2 compares the average runtime of the bonnet model from [17] and the modified bonnet model.

**Table 2.** Average Runtime of the Models

| Model | Input | Hardware | FPS |
|---|---|---|---|
| Bonnet Model [17] | All (14 Channels) | Intel i7+GTX 1080Ti | 22.7 |
| Modified Bonnet Model (This Work) | All (10 Channels) | Intel i7 + NVIDIA 940 MX | **2.5** |

It is difficult to get a direct comparison between our modified 10-channel Bonnet and the 14-channel Bonnet as the latter was trained on 3 datasets- Bonn Stuttgart and Zurich whereas we were able to train it only on the Bonn dataset due to the lack of availability of other two datasets. However, we have tabulated our results and tried to compare them with [17] in order to check and validate the performance. In terms of mean accuracy, our model achieved an accuracy of 99.47% as compared to model performance of 94.74% from [17] on the Bonn Dataset. Our model performs at a mean IOU of 98.03% as compared to the model performance from [17] which is 80.8%. In terms of precision and recall, the model performs quite similarly for the crop and soil class. While the performance in terms of weed class is not up to mark as obtained in [17]. However, it performs quite similarly to ground truth on both Bonn and CWFID datasets. Training the network on more labeled data would definitely help increase the classification performance further.

As the hardware which is used for the deployment is different, a direct comparison in terms of FPS cannot be made. The modified Bonnet Model achieved an average latency of 2.5 fps on Intel i7 + NVIDIA 940 MX.

The performance of the modified Bonnet Model on the Bonn [19] and CWFID [20] Datasets are depicted in the Fig. 17. and Fig 18 respectively. Here the first image from the left is the input feed from the camera. The soil, crop and weed classes have been assigned with different colour scheme namely blue, green, and red respectively. The remaining images depict the correct label, class-wise prediction, and finally the model prediction.

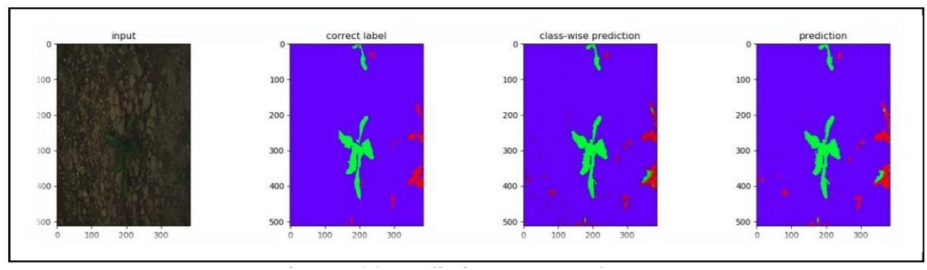

**Fig. 17.** Prediction on Bonn dataset.

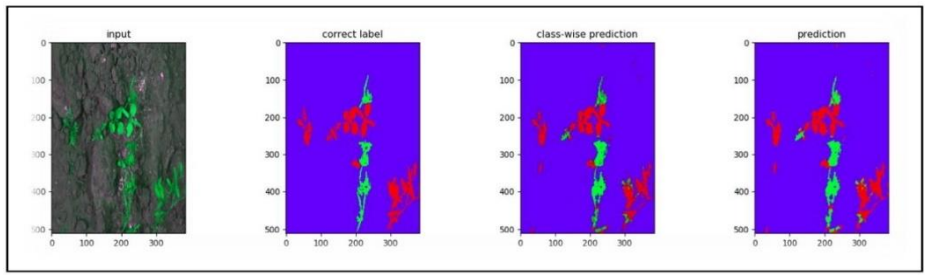

**Fig. 18.** Prediction on CWFID dataset.

Here in Fig. 19, the left window refers to the video feed from the camera in the Gazebo simulation of our AGRIBOT. And the right window is the model prediction of the crop-weed classification model on that camera feed. Fig. 20 shows the prediction on the images obtained from the nearby fields. The blue, green and red colour depicts soil, crop and weed respectively.

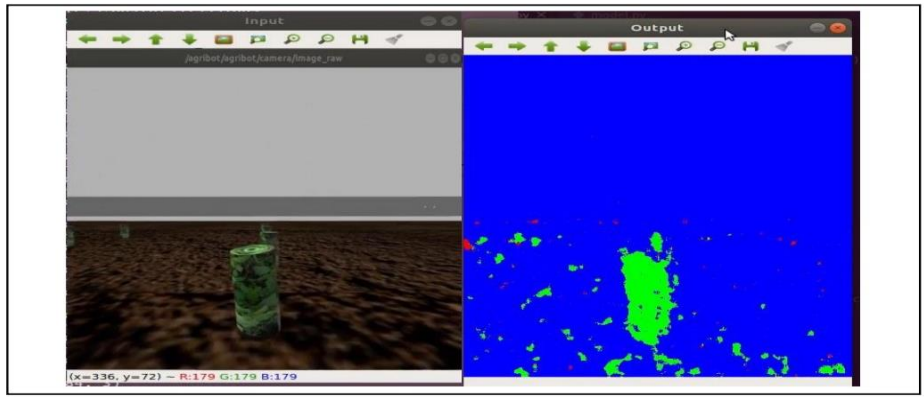

**Fig. 19.** Model Prediction in the Gazebo Simulator.

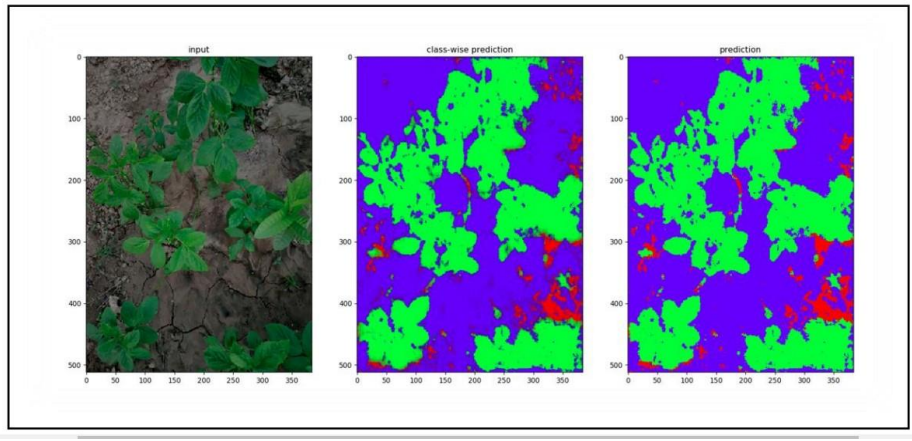

**Fig. 20.** Predicted Image from the nearby field

From the above predictions, it can be observed that the model has now correctly learned to recognize the vegetation mask in the image. It has also tried to classify features like less green or dry parts of plants as weeds. The model has started to learn to classify weeds and the predictions can still be improved by providing more data to it.

## 6      Conclusion

This paper presents an idea of making an agriculture robot that is specifically used for the detection of weeds. The hardware consisting of both mechanical and electronics systems was proposed by considering various key parameters. Data from various sensor plugins was acquired and processed using various filtering and sensor fusion techniques to get important information like distance and angle to the destination which is used for autonomous navigation of the robot. Simultaneous control of Linear and Rotational motion was done on the robot which was used to reach the destination.

Semantic Segmentation approach was best suited for the agriculture robot. Two Semantic Segmentation architectures namely UNet and Modified Bonnet were trained and tested on 2 different datasets: CWFID and Bonn dataset. Modified Bonnet Model was selected over Unet due to its better metrics and performance and lesser number of parameters which makes it possible to run on a real-time for our application. It model achieved a mean accuracy of 99.47%, a mean IoU of 98.03% and loss of 0.00348 units on the Bonn Dataset. The proposed model has an average latency of 2.5 fps on i7 + NVIDIA 940 MX. This makes it possible to deploy the model on an on-board processor like NVIDIA Jetson Nano for real-world application.

For improving the autonomous navigation and making the trajectory of the AGRIBOT more smooth, various other techniques such as DGPS (Differential GPS), implementation of predictive filters and control techniques can be tested. For minimizing crop damage, data from the camera can be fused and used for precise navigation of the robot. For the crop-weed classification task, training the network over more labeled data of different crops and vegetation would significantly improve the performance across all the classes and generalize it well to identify different weed species. The design of the AGRIBOT was experimented and validated on the Gazebo simulator.